\pdfoutput=1
\documentclass{article}

    \usepackage[nonatbib,preprint]{neurips_2024}
    \usepackage[numbers]{natbib}

\usepackage[utf8]{inputenc} % allow utf-8 input
\usepackage[T1]{fontenc}    % use 8-bit T1 fonts
\usepackage{hyperref}       % hyperlinks
\usepackage{url}            % simple URL typesetting
\usepackage{booktabs}       % professional-quality tables
\usepackage{amsfonts}       % blackboard math symbols
\usepackage{nicefrac}       % compact symbols for 1/2, etc.
\usepackage{microtype}      % microtypography
\usepackage{xcolor}         % colors
\usepackage{amsmath}
\usepackage{amssymb}
\usepackage{mathtools}
\usepackage{amsthm}
\usepackage{listings}
\usepackage{bbm}

\usepackage{amsmath,amsfonts,bm}

\def\eqref#1{equation~\ref{#1}}
\def\1{\bm{1}}

\def\rmX{{\mathbf{X}}}

\def\vk{{\bm{k}}}

\def\vm{{\bm{m}}}

\def\vq{{\bm{q}}}

\def\vv{{\bm{v}}}

\def\vx{{\bm{x}}}

\DeclareMathAlphabet{\mathsfit}{\encodingdefault}{\sfdefault}{m}{sl}
\SetMathAlphabet{\mathsfit}{bold}{\encodingdefault}{\sfdefault}{bx}{n}

\newcommand{\softmax}{\mathrm{Softmax}}

\DeclareMathOperator{\defeq}{\stackrel{\text{def}}{\; = \;}}
\DeclareMathOperator{\diag}{\text{diag}}

\theoremstyle{plain}
\newtheorem{theorem}{Theorem}[section]
\newtheorem{proposition}[theorem]{Proposition}

\theoremstyle{definition}
\newtheorem{definition}[theorem]{Definition}
\newtheorem{assumption}[theorem]{Assumption}
\theoremstyle{remark}

\title{Bridging Associative Memory and \\Probabilistic Modeling}

\author{%
  Rylan Schaeffer\thanks{Correspondence to \texttt{rschaef@cs.stanford.edu} and \texttt{sanmi@cs.stanford.edu}.} \\
  Stanford CS\\
  \And
  Nika Zahedi \\
  Stanford EE \\
  \And
  Mikail Khona\\
  MIT Physics\\
  \And
  Dhruv Pai\\
  Stanford CS\\
  \And
  Sang T. Truong\\
  Stanford CS\\
  \AND
  Yilun DU\\
  MIT EECS\\
  \And
  Mitchell Ostrow\\
  MIT BCS\\
  \And
  Sarthak Chandra\\
  MIT BCS\\
  \And
  Andres Carranza\\
  Stanford CS\\
  \AND
  Ila R Fiete\\
  MIT BCS\\
  \And
  Andrey Gromov\\
  UMD Physics\\
  \And
  Sanmi Koyejo$^*$\\
  Stanford CS
}

\begin{document}

\maketitle

\begin{abstract}
Associative memory and probabilistic modeling are two fundamental topics in artificial intelligence. The first studies recurrent neural networks designed to denoise, complete and retrieve data, whereas the second studies learning and sampling from probability distributions.
Based on the observation that associative memory's energy functions can be seen as probabilistic modeling's negative log likelihoods, we build a bridge between the two that enables useful flow of ideas in both directions. We showcase four examples:
First, we propose new energy-based models that flexibly adapt their energy functions to new in-context datasets, an approach we term \textit{in-context learning of energy functions}.
Second, we propose two new associative memory models: one that dynamically creates new memories as necessitated by the training data using Bayesian nonparametrics, and another that explicitly computes proportional memory assignments using the evidence lower bound.
Third, using tools from associative memory, we analytically and numerically characterize the memory capacity of Gaussian kernel density estimators, a widespread tool in probabilistic modeling. %, as a function of number of data points, data dimensionality, choice of kernel and kernel bandwidth. 
Fourth, we study a widespread implementation choice in transformers -- normalization followed by self attention -- to show it performs clustering on the hypersphere.
% XXX unclear from the sentence above how the fourth contribition is connected to associative memory or probabilistic modeling, in the way that the other contributions clear are. 
Altogether, this work urges further exchange of useful ideas between these two continents of artificial intelligence.
\end{abstract}
\section{Introduction}

Associative memory concerns dynamical systems with state $\vx(t) \in \mathbb{R}^D$ and dynamics $f: \mathcal{X} \times \Theta \rightarrow \mathcal{X}$ constructed so that the dynamics denoise, complete and/or retrieve training data:
\begin{equation}
    \tau \frac{d}{dt}\vx(t) \defeq f_{\theta}(\vx(t)),
    \label{eqn:associative_memory_dynamics_recurrent_network}
\end{equation}
 
Associative memory research is often interested in the stability and capacity of memory models, e.g.,  \cite{hopfield1982neural, hopfield1984neurons, hopfield1986computing,tanaka1980analytic,abu1985information,crisanti1986saturation,mceliece1987capacity,torres2002storage,folli2017maximum,sharma2022content}, questions that were often answered by showing the dynamics monotonically non-increase energy functions $E_{\theta}(x)$; recent work then introduced ``modern" associative memory that explicitly define dynamics as minimizing energy functions \cite{krotov2016dense,demircigil2017model,barra2018new,ramsauer2020hopfield,krotov2020large}:
\begin{equation}
    \tau \frac{d}{dt}\vx(t) \defeq - \nabla_{\vx} \, E_{\theta}(\vx(t)).
    \label{eqn:associative_memory_dynamics_energy}
\end{equation}

By doing so, a bridge was constructed to probablistic modeling. Probabilistic modeling often aims to learn a probability distribution $p_{\theta}(\vx)$ with parameters $\theta$ using training dataset $\mathcal{D} \defeq \{ \vx_n \}_{n=1}^N$, which can be expressed in Boltzmann distribution form \cite{bishop2006pattern}:
\begin{equation}
    p_{\theta}(\vx) = \frac{\exp \big(-E_{\theta}(\vx) \big)}{ Z_{\theta}} \quad \Rightarrow \quad - \nabla_{\vx} E_{\theta}(\vx) = \nabla_{\vx} \log p_{\theta}(\vx),
    \label{eqn:boltzmann_distribution}
\end{equation}
where $Z(\theta) \defeq \int_{\vx \in \mathcal{X}} \exp(-E(\vx)) \, d\vx$ is the partition function and the energy's negative derivative is the so-called score function.
This connection - that an associative memory's recurrent dynamics can be seen as performing gradient descent on the negative log likelihood or that performing gradient descent on the negative log likelihood can be seen as creating a dynamical system minimizing an energy functional - has indeed been noted many times before \cite{barra2012equivalence,scellier2017equilibrium,radhakrishnan2018memorization, radhakrishnan2020overparameterized,fuentes2019modal,annabi2022relationship,hoover2023memory,ambrogioni2023search}
However, prior work often focused on particular settings, missing the forest for the trees.
In this work, we aim to prominently highlight this relationship and show how it can more generally drive a meaningful exchange of ideas in both directions.
Our specific contributions include:
\begin{enumerate}
    \item Inspired by the capability of associative memory models to flexibly create new energy landscapes for new datasets, we propose a new probabilistic energy-based model (EBM) that can similarly easily adapt their computed energy landscapes based on in-context data \textit{without modifying their parameters}. Due the spiritual similarity of this capability with in-context learning of transformer-based language models, we term this \textbf{in-context learning of energy functions}. To the best of our knowledge, this is the first instance of in-context learning with transformers \textit{where the output space differs from the input space}.
    \item We identify how recent research in the associative memory literature corresponds to learning memories for fixed energy functional forms and propose two new associative memory models originating in probabilistic modeling: The first enables creating new memories as necessitated by the data by leveraging Bayesian nonparametrics, while the second enables computing cluster assignments using the evidence lower bound.
    \item We demonstrate that kernel density estimators (KDEs), a widely used probabilistic method, have memory capacities (i.e., a maximum number of memories that can be successfully retrieved), and analytically and numerically characterize capacity, retrieval and failure behaviors of Gaussian KDEs.
    \item We mathematically show that a widely-employed implementation decision in modern transformers -- normalization before self-attention -- approximates clustering on the hypersphere using a mixture of inhomogeneous von Mises-Fisher distributions, as has been conjectured before and observed numerically \cite{liu2023deja, geshkovski2024emergence}. Further, we provide a theoretical ground for recent normalization layers in self-attention that have shown to bestow stability to transformer training dynamics \cite{dehghani2023scaling,wortsman2023small}.
\end{enumerate}

\section{In-Context Learning of Energy Functions}

\paragraph{Motivation for In-Context Learning of Energy Functions}

One useful property of associative memory is their flexibility: the memories (i.e., training data) $\mathcal{D} \defeq \{\vx_n\}_{n=1}^N$ can be hot-swapped to immediately change the energy landscape. For examples, the Hopfield Network \cite{hopfield1982neural} has energy:
\begin{equation}
    E_{\theta}^{HN}(\vx) \defeq -\frac{1}{2} \vx^T \Big( \frac{1}{N}\sum_n \vx_n \vx_n^T \Big) \vx
    \label{eqn:hopfield_network_energy}
\end{equation}
and the Modern Continuous Hopfield Network (MCHN) \cite{ramsauer2020hopfield, krotov2020large} has energy \footnote{We omit terms constant in $\vx$ because they do not affect the fixed points of the energy landscape.}:
\begin{equation}
    E_{\theta}^{MCHN}(\vx) \defeq - \frac{1}{\beta}\log \Bigg(\sum_n \exp \big(\beta \vx^T \vx_n \big) \Bigg) + \frac{1}{2} \vx^T \vx,
    \label{eqn:modern_continuous_hopfield_network_energy}
\end{equation}
%
%where parameters $\theta$ are dataset $\mathcal{D}$ and the inverse temperature $\beta > 0$.
In both examples, if the dataset $\mathcal{D}$ is replaced with a different dataset $\mathcal{D}'$, the energy landscape immediately adjusts.
In contrast, in probabilistic modeling, energy-based models (EBMs) typically have no equivalent capability because the learned energy $E_{\theta}(\vx)$ depends on pretraining data $\mathcal{D}$ only through the learned neural network parameters $\theta = \theta(\mathcal{D})$ \cite{du2019implicit,nijkamp2020anatomy,du2020compositional,du2020improved, du2021unsupervised}.
However, there is no fundamental reason why EBMs cannot be extended to be conditioned on entire datasets as associative memory models often are, and we thus demonstrate how to endow EBMs with such capabilities.

\begin{figure*}
    % \centering
    % https://wandb.ai/rylan/icl-ebm/runs/gl2alcnw?workspace=user-rylan
    \includegraphics[width=1.\textwidth]{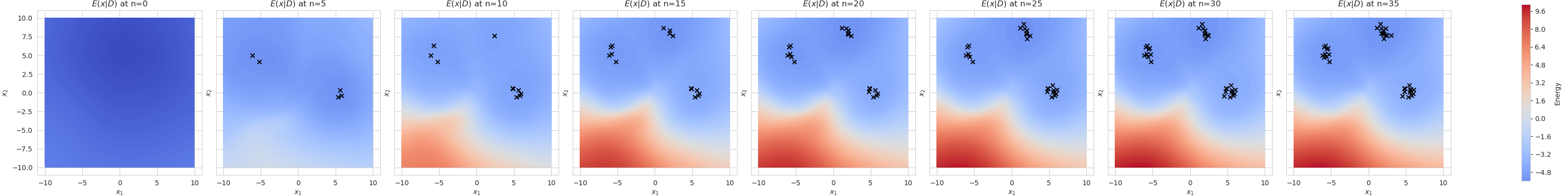}
    \includegraphics[width=1.\textwidth]{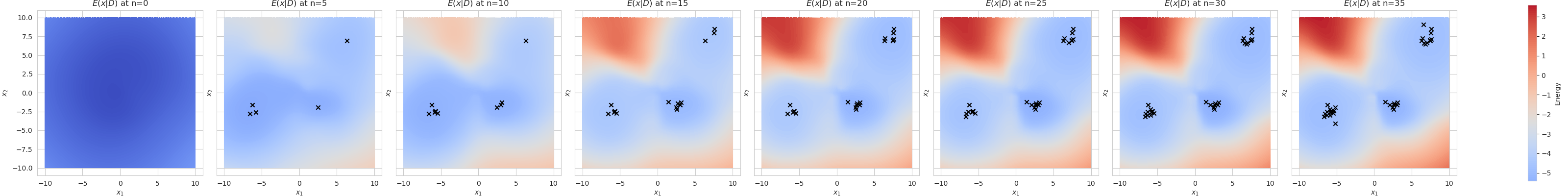}
    \includegraphics[width=1.\textwidth]{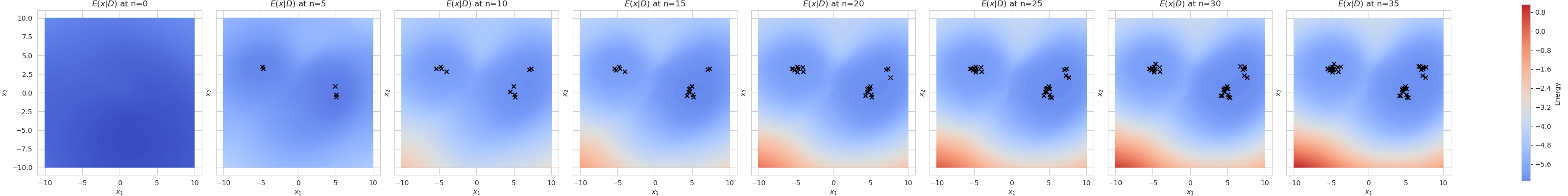}
    \includegraphics[width=1.\textwidth]{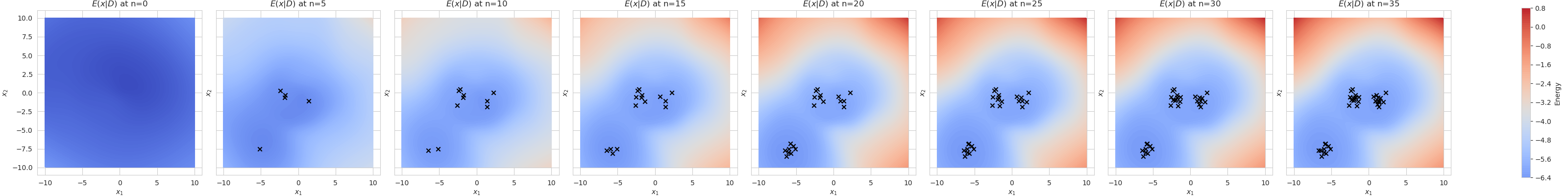}
    \caption{\textbf{In-Context Learning of Energy Functions.} Transformers learn to compute energy functions $E_{\theta}^{ICL}(x|\mathcal{D})$ corresponding to probability distributions $p_{\theta}^{ICL}(x|\mathcal{D})$, where $\mathcal{D}$ are in-context datasets that vary during pretraining. At inference, when conditioned on a new in-context dataset, the transformer computes a new energy function using fixed parameters $\theta$. Left-to-Right: The transformers' energy landscapes sharpen as additional in-context data are conditioned upon.}
    \label{fig:icl_ebm}
\end{figure*}

\paragraph{Learning In-Context Energy Functions}

We therefore propose energy-based modeling of dataset-conditioned distributions. This EBM should accept as input an arbitrarily sized dataset $\mathcal{D}$ and a single datum $\vx$, and adaptively change its output energy function $E_{\theta}^{ICL}(\vx|\mathcal{D})$ based on the input dataset $\mathcal{D}$ \textit{without changing its parameters} $\theta$. This corresponds to learning the conditional distribution:
\begin{align}
    p_{\theta}^{ICL}(\vx|\mathcal{D}) = \frac{\exp\big( -E_{\theta}^{ICL}(\vx|\mathcal{D}) \Big)}{Z_{\theta}(\mathcal{D})}
\end{align}
Based on a similarity to in-context learning capabilities of language models \cite{brown2020language}, we call this \textbf{\textit{in-context learning of energy functions}} (ICL-EBM). To implement this, we use a causal transformer with a GPT-like architecture \cite{vaswani2017attention, radford2018improving, radford2019language} that replaces the conditional distribution $p(x_n|x_{<n})$ at each index $n$ with its corresponding energy function $E(x_n|x_{<n})$; see App. \ref{app:sec:implementation_details_icl_ebm} for implementation details.
This means that the transformer outputs a scalar variable at every index: $E(x_2|x_1), E(x_3|x_2, x_1), E(x_4|x_3,x_2, x_1), \dots$. This scalar at each index is the model’s estimate of the energy at the last sample ($n^{\text{th}}$) input data point, assuming an energy function constructed by the previous $n-1$ datapoints. The transformer is trained to minimize the negative log conditional probability, averaging over all possible in-context datasets:
\begin{equation}
    \mathcal{L}(\theta) \; \defeq \; \mathbb{E}_{p_{data}} \Bigg[ \mathbb{E}_{\vx, \mathcal{D}\sim p_{data}} \Big[- \log p_{\theta}^{ICL}(\vx | \mathcal{D}) \Big] \Bigg].
    \label{eqn:icl_energy_loss_fn}
\end{equation}
Due to the intractable partition function in Eqn. \ref{eqn:icl_energy_loss_fn}, we minimize the loss using contrastive divergence \cite{hinton2002training}. Letting $\vx^{+}$ denote real training data and $\vx^{-}$ denote confabulatory data sampled from the learned energy function, the gradient of the loss function is given by:
\begin{equation*}
\begin{aligned}
    &\nabla_{\theta} \mathcal{L}(\theta) = \nabla_{\theta} \, \mathbb{E}_{p_{data}} \Bigg[ \mathbb{E}_{\vx^{+} \mathcal{D}\sim p_{data}} \Big[- \log p_{\theta}(\vx | \mathcal{D}) \Big] \Bigg]\\
    % &= \mathbb{E}_{p_{data}} \Bigg[ \mathbb{E}_{x, \mathcal{D}\sim p_{data}} \Big[\nabla_{\theta} E_{\theta}^{ICL}(x, \mathcal{D}) \Big] \Bigg]\\
    % &\quad \quad + \mathbb{E}_{p_{data}} \Bigg[ \mathbb{E}_{\mathcal{D}\sim p_{data}} \Big[\nabla_{\theta} \log Z_{\theta}^{ICL}(\mathcal{D}) \Big] \Bigg]\\
    &= \mathbb{E}_{p_{data}} \Bigg[ \mathbb{E}_{\vx^{+} | \mathcal{D}\sim p_{data}} \Big[\nabla_{\theta} E_{\theta}^{ICL}(\vx^{+}, \mathcal{D}) \Big] - \mathbb{E}_{\mathcal{D}\sim p_{data}} \Big[\mathbb{E}_{\vx^{-} \sim p_{\theta}^{ICL}(\vx | \mathcal{D})} \big[ \nabla_{\theta} E_{\theta}^{ICL}(\vx^{-} | \mathcal{D}) \big] \Big] \Bigg].
\end{aligned}
\end{equation*}

\paragraph{Sampling From In-Context Energy Functions}

To sample from the conditional distribution $p_{\theta}(\vx|\mathcal{D})$, we follow standard practice \cite{hinton2002training,du2019implicit,du2020improved}: We first choose $N$ data (deterministically or stochastically) to condition on, and sample $\vx_0^{-} \sim \mathcal{U}$ for some $\mathcal{U}$ to compute the initial energy $E_{\theta}(\vx_0^{-}| \mathcal{D})$. We then use Langevin dynamics to iteratively increase the probability of $\vx_0^{-}$ by sampling with $\omega_t \sim \mathcal{N}(0, \sigma^2)$ and minimizing the energy with respect to $\vx_t^{-}$ for $t=[T]$ steps:
\begin{equation}
    \vx_{t+1}^{-} \leftarrow \vx_{t}^{-} - \alpha \nabla_{\vx} \, E_{\theta}^{ICL}(\vx_{t}^{-}| \mathcal{D}) + \omega_{t}.
\end{equation}
This in-context learning of energy functions is akin to Mordatch et. al (2018)\citep{mordatch2018concept}, but rather than conditioning on a ``mask" and ``concepts", we instead condition on sequences of data from the same distribution and we additionally replace the all-to-all relational network with a causal transformer.

\paragraph{Experiments for In-Context Learning of Energy Functions}

As proof of concept, we train causal transformer-based ICL-EBMs on synthetic datasets. The transformers have $6$ layers, $8$ heads, $128$ embedding dimensions, and GeLU nonlinearities \cite{hendrycks2016gaussian}. The transformers are pretrained on a set of randomly sampled synthetic 2-dimensional mixture of three Gaussians with uniform mixing proportions with Langevin noise scale $0.01$ and 15 MCMC steps of size $\alpha = 3.16$. After pretraining, we then freeze the ICL-EBMs' parameters and measure whether the model can adapt its energy function to new in-context datasets drawn from the same distribution as the pretraining datasets. The energy landscapes of frozen ICL EBMs display clear signs of in-context learning (Fig. \ref{fig:icl_ebm}). To the best of our knowledge, \textit{this is the first instance of in-context learning where the input and output spaces differ}, in stark comparison with more common examples of in-context learning such as language modeling \cite{brown2020language}, linear regression \cite{garg2022can} and image classification \cite{chan2022data}.% Transformers are more capable than previously known.
\section{Learning Memories for Associative Memory Models}
\label{sec:learning_memories_for_associative_memory}

\paragraph{Connecting Research on Learning Memories}

In many associative memory models, the energy functions are defined a priori. However, one might instead \textit{learn} an energy function.
One approach to do so is to transform each datum $\vx_n$ into a learnt representation $\bm \xi_n$ that is then evolved through a classical energy landscape \cite{ramsauer2020hopfield,hoover2023energytransformer}.
%The energy transformer (ET) evolves masked datapoints through the joint energy landscape of a transformer and Hopfield network until convergence at some fixed points. For instance. The memories are embedded explicitly as learned weights in the network in the ET.
A complementary approach is to learn $K$ memories using $N$ data, an approach recently taken by Saha et. al (2023) \citep{saha2023end} called \textbf{Cl}ustering with \textbf{A}ssociative \textbf{M}emories (ClAM).
We show how ClAM is closely connected to probabilistic modeling; by making the connection explicit, we then propose two new associative memory models 
(Sec. \ref{subsec:latent_var_energy_functions}, \ref{subsec:nonparametric_energy_functions}) as well as a combined form (Sec. \ref{subsec:nonparametric_latent_variable_energy_function}). ClAM's energy is:
\begin{equation}
    E_{\theta}^{ClAM}(\vx) \defeq -\frac{1}{\beta} \log \Bigg(\sum_k \exp \big(-\beta ||\bm\mu_k - \vx||^2 \big)  \Bigg),
    \label{eqn:saha_2023_energy}
\end{equation}
where parameters $\theta$ are learnable memories $\{\bm \mu_k \}_{k=1}^K$ and inverse temperature $\beta$. Its dynamics are:
\begin{align}
    \tau \frac{d\vx(t)}{dt}
    &= \sum_k (\bm\mu_k - \vx) \; \softmax \Big(-\beta ||\bm\mu_k - \vx||^2 \Big).
    \label{eqn:saha_2023_dynamics}
\end{align}

To learn the memories $\{\bm \mu_k \}_k$, ClAM perform gradient descent on the reconstruction loss:
\begin{equation}
    \mathcal{L}^{ClAM}\Big(\{\bm \mu_k \}_k \Big) \defeq \sum_{n=1}^N \Big|\Big|\vx_n - \vx_n^{\{\bm \mu_k\}}(T) \Big|\Big|^2,
    \label{eqn:saha_2023_loss}
\end{equation}
where $\vx_n^{\{\bm \mu_k\}}(T)$ is the state of the AM network with memories $\{\bm \mu_k\}_{k=1}^K$ having been initialized at $\vx(0) = \vx_n$ and then following the dynamics for $T$ time.
This associative memory model has a spiritual connection to probabilistic modeling's finite Gaussian mixture model with homogeneous isotropic covariances $\Sigma_K = 2 \beta^{-1} I_D$ and uniform mixing proportions $\pi_k = 1/K$:
\begin{equation*}
    p_{\theta}^{ClAM}(\vx) = \sum_{k} \mathcal{N}(\vx ;\bm\mu_k, \Sigma_k) \, \pi_k.
\end{equation*}
%
% We also see that the fixed points of the associative memory dynamics are the memories weighted by the cluster assignment posteriors; that is, if the network is initialized at $\vx(0) = \vx$, then $\vx^* \defeq \sum_k p(z=k|\vx;\theta) \bm \mu_k$ is a fixed point:
% %
% \begin{align*}
%     \tau \frac{d\vx^*}{dt} &= \Bigg( \underbrace{\sum_k \bm \mu_k \softmax \big(-\beta ||\bm\mu_k - \vx^*||^2 \big)}_{\defeq \vx^*} -  \vx^* \underbrace{\sum_k \softmax \big(-\beta ||\bm\mu_k - \vx^*||^2 \big)}_{=1} \Bigg) = 0.
% \end{align*}

\begin{figure}
    \centering
    \includegraphics[width=0.7\linewidth]{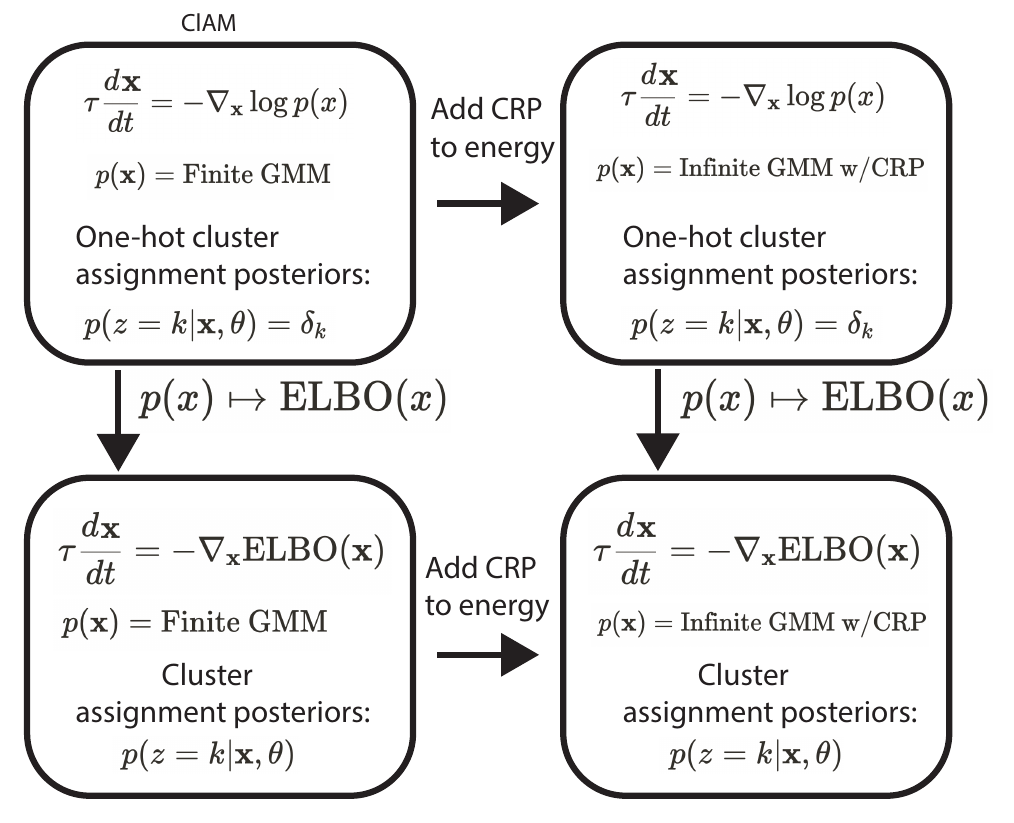}
    \caption{\textbf{New Associative Memory Models: Latent Variable and Bayesian Nonparametric.} We propose two new associative memory models that can compute proportional cluster assignments using the evidence lower bound (top to bottom) and can create new memories using Bayesian nonparametrics (left to right). Applying both together results in an associative memory model capable of creating new memories and simultaneously explicitly computing cluster assignment posteriors.
    }
    \label{fig:ClAM_schematic}
\end{figure}

Choosing non-uniform mixing proportions corresponds to ClAM's ``weighted clustering," and choosing a von Mises-Fisher likelihood corresponds to their ``spherical clustering"; one can, of course, choose other likelihoods e.g. Laplace, uniform, L\'evy, etc.
In the language of probabilistic modeling, ClAM is akin to ``Generalized Expectation Maximization (EM)" \cite{dempster1977maximum,xu1996convergence, neal1998view, salakhutdinov2003optimization} applied to a mixture model. Generalized EM's two alternating phases morally correspond to ClAM's two alternating phases. Generalized EM's expectation step prescribes increasing the log likelihood with respect to the cluster assignment posterior probabilities, which corresponds to ClAM minimizing its energy function (Eqn. \ref{eqn:saha_2023_energy}) with respect to the particle $\vx(t)$ by rolling out the dynamics (Eqn. \ref{eqn:saha_2023_dynamics}).
Generalized EM's maximization step, which maximizes the log-likelihood with respect to the parameters $\theta$, mirrors ClAM's shaping of the energy landscape by taking a gradient step with respect to the parameters $\theta$.
% Thus, ClAM is a dynamical system whose forward dynamics cluster individual data. This is similar to recent nonlinear control work \cite{romero2019convergence,chatterjee2022analysis}, but differs in that ClAM updates the parameters $\theta$ via backpropagation \cite{rumelhart1986learning} rather than in its forward dynamics. By making this connection, we can now propose two new classes of associative memory models: latent variable and Bayesian nonparametric.

\paragraph{Latent Variable Associative Memory Models}
\label{subsec:latent_var_energy_functions}

One limitation of ClAM's associative memory is that, in the context of clustering, it provides no mechanism to obtain the cluster assignment posteriors $p_{\theta}(z=k|\vx;\theta)$. Such posteriors are useful for probabilistic uncertainty quantification and also for designing more powerful associative memory networks (Sec. \ref{subsec:nonparametric_energy_functions}). We propose a new associative memory model that preserves the fixed points and their stability properties but  computes the cluster assignment posteriors explicitly by converting the evidence lower bound (ELBO) -- a widely used lower bound in probabilistic modeling -- into an energy function. Recall that the log likelihood can be lower bounded by Jensen's inequality:
\begin{align*}
    \log p_{\theta}^{ClAM}(\vx)
    % \log \sum_k p(\vx, z=k; \theta)\\
    % &= \log \sum_k q(z=k) \frac{p(\vx, z; \theta)}{q(z=k)}\\
    % &\geq \sum_z q(z=k) \log \frac{p(\vx, z = k; \theta)}{q(z=k)}\\
    & \geq \mathbb{E}_{q(z)}[\log p_{\theta}(\vx, z=k)] + H[q(z)],
\end{align*}
where $H(\cdot)$ is the entropy. Denote $q(z)$ with the probability vector $\vq \in \Delta^{K-1}$ and define the energy:
\begin{align*}
    E_{\theta}^{ClAM+ELBO}(\vq) &\defeq %-\mathbb{E}_{q(z)}[\log p(\vx, z=k; \theta)] - H[q(z)]\\
    -\sum_{k=1}^K \vq_k \log p_{\theta}(\vx, z=k) + H(\vq)
    \label{eqn:ClAM_elbo_energy}
\end{align*}

To ensure that $\vq(t)$ remains a probability vector, we reparameterize $\vq(t)$ using $\vv(t) \in \mathbb{R}^K$ with $\vq(t) = \softmax(\vv(t))$. This yields an associative memory model where the state $\vv(t)$ lives in the number-of-clusters-dimensional logit space $\mathbb{R}^K$ rather than data space $\mathcal{X}$. 
Recalling that the gradient of probability vector $\vq$ with respect to its logits $\vv$ can be expressed in matrix notation as $\nabla_{\vv} \vq = \diag(\vq) - \vq \vq^T \in \mathbb{R}^{K \times K}$, the dynamics in logit space are:
\begin{equation}
    \tau \frac{d}{dt} \vv(t) \defeq - \nabla_{\vv} E_{\theta}^{ClAM+ELBO}(\vq(\vv(t))) = \Big( \diag(\vq) - \vq \vq^T \Big) \Big( \log p_{\theta}(\vx, z) - \log \vq - \bm 1 \Big)
    \label{eqn:ClAM_elbo_dynamics}
\end{equation}

% In $\vq$ space, these dynamics have a single fixed point corresponding to the cluster assignment posteriors: $\vq^* = p(z=k|\vx;\theta)$. However 
Due to the invariance of Softmax to constant offsets, the dynamics do not have a single fixed point but rather an invariant set in $\vv$ space: $\softmax(\vv + c)_k = (\exp \vv_k \exp c) / (\sum_i \exp \vv_i \exp c) = \exp \vv_k / \sum_i \exp \vv_i =\softmax(\vv)_k$. This implies the same symmetry exists in the energy function, $E_{\theta}^{ClAM+ELBO}(\vv + c) = E_{\theta}^{ClAM+ELBO}(\vv)$, thus all minima $\vv^*$ (the fixed points of the energy function) are in fact invariant sets $\vv^* + \alpha \mathbf{1}$, with $\alpha \in \mathbb{R}$.
Like ClAM, convergence to a local minimum is guaranteed because the energy is monotonically non-increasing: 
\begin{align*}
\frac{d}{dt}E(\vq(\vv(t))) = \nabla_{\vv} E^{ClAM+ELBO}(\vq(\vv(t))) \cdot \frac{d}{dt}\vv(t) = -||\nabla_{\vv} E(\vq(\vv(t)))||^2 \leq 0
\end{align*}

\begin{figure*}
    \centering
    \includegraphics[width=\textwidth]{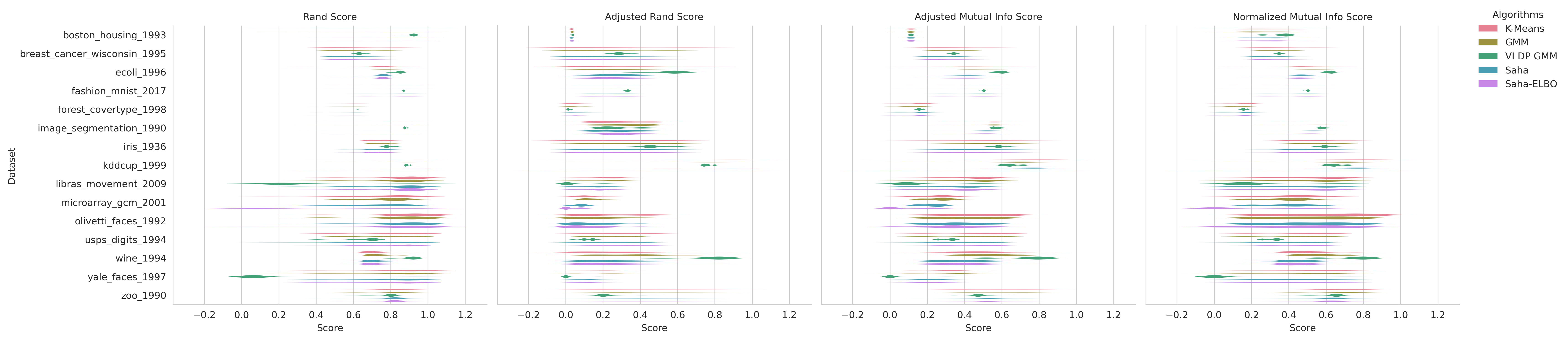}
    \caption{\textbf{ClAM, ClAM+ELBO, and various baselines' performance on supervised metrics for standard benchmark datasets.} ClAM+ELBO is competitive with ClAM across benchmark tasks in supervised metrics.}
    \label{fig:ClAM_dual_benchmark_performance_supervised_metrics}
\end{figure*}

\begin{figure*}
    \centering
    \includegraphics[width=\textwidth]{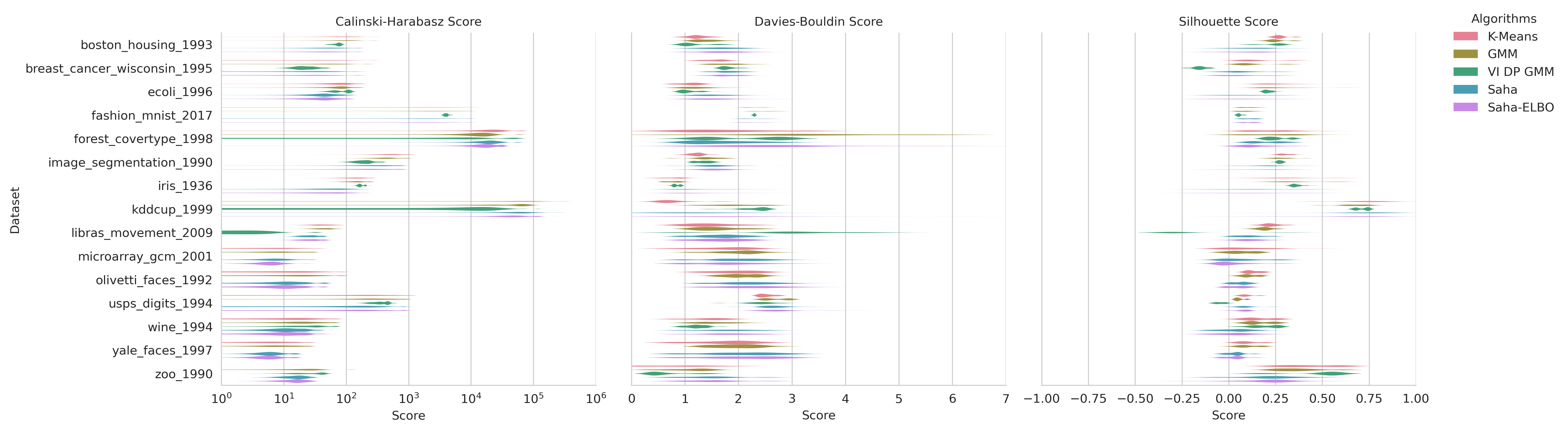}
    \caption{\textbf{ClAM, ClAM+ELBO, and various baselines' performance on unsupervised metrics for standard benchmark datasets.} ClAM+ELBO is competitive with ClAM across benchmark tasks in unsupervised metrics.}
    \label{fig:ClAM_dual_benchmark_performance_unsupervised_metrics}
\end{figure*}

Empirically, we find that ClAM-ELBO is competitive with ClAM across a wide range of benchmarks under both supervised and unsupervised metrics (Fig. \ref{fig:ClAM_dual_benchmark_performance_supervised_metrics}, Fig. \ref{fig:ClAM_dual_benchmark_performance_unsupervised_metrics}). 

\paragraph{Bayesian Nonparametric Associative Memory Models}
\label{subsec:nonparametric_energy_functions}

Based on the connection to probabilistic modeling, one can also construct associative memory models that learn the number of memories as necessitated by the data. This is interesting biologically and computationally: biologically, animals create new memories throughout their lives, and computationally, choosing the right number of clusters in clustering is a perennial problem \cite{thorndike1953belongs, rousseeuw1987silhouettes,bischof1999mdl, pelleg2000x, tibshirani2001estimating, sugar2003finding, hamerly2003learning, kulis2012revisiting}.

To create an AM network with the ability to create new memories, we propose leveraging Bayesian nonparametrics based on combinatorial stochastic processes \cite{pitman2006combinatorial}. Specifically, we will use the Chinese Restaurant Process (CRP) \cite{ blackwell1973ferguson, antoniak1974mixtures,aldous1985exchangeability,teh2010dirichlet}\footnote{The 1-parameter $CRP(\alpha, d=0)$ and the 2-parameter $CRP(\alpha, d)$ correspond to the Dirichlet Process and the Pitman-Yor Process, respectively.}. The CRP defines a probability distribution over partitions of a set that can then be used as a  prior over the number of clusters as well as a prior over the number of data per cluster. Specifically, let $\alpha > 0, d \in [0, 1)$ be hyperparameters and $K_{<n} \defeq \max \{z_1, ..., z_{n-1}\}$ denote the number of clusters after the first $n-1$ data. Then $CRP(\alpha, d)$ defines a conditional prior distribution on cluster assignments:
\begin{align*}
    p(z_n = k | z_{<n}, \alpha, d) \; \defeq &\frac{1}{n - 1 + \alpha}
    \begin{cases}
    - d + \sum_{n' < n}   \mathbbm{I}(z_{n'} = k) & \text{ if } 1 \leq k \leq K_{<n}^+\\
    \alpha + d \cdot K_{<n}^+ & \text{ if } k = K_{<n}^+ + 1\\
    0 & \text{otherwise}
    \end{cases}
    \label{eqn:crp_conditional_definition}
\end{align*}
The hyperparameter $\alpha > 0$ controls how quickly new clusters form, and the hyperparameter $d \in [0, 1)$ controls how quickly existing memories accumulate mass. We propose using the CRP to define a novel associative memory model that creates new memories. Let $\theta$ denote the model parameters: $K^+$ is the number of clusters, $\{\tilde{\pi}_k\}_{k=1}^{K^+}$ are the number of data assigned to each existing cluster, and $\{\bm\mu_k, \Sigma_k \}_{k=1}^{K^+}$ are the means and covariances of the clusters.
Then, assuming an isotropic Gaussian likelihood $\Sigma_k = 2\beta^{-1} I_D$ and assuming an isotropic Gaussian prior on the cluster means $\bm \mu_k \sim \mathcal{N}(\bm 0, 2\rho^{-1} I_D)$, the probability of datum $\vx$ is:
\begin{align*}
    p_{\theta}^{ClAM+CRP}(\vx) 
     \; \defeq \;p(\vx | z = K^{+} + 1;\theta) \, p(z = K^{+} + 1;\theta) + \sum_{k=1}^{K^+} p(\vx | z = k;\theta) \, p(z = k;\theta)
\end{align*}
Using the same process as before, we can convert the probability distribution into an energy function via the inverse temperature-scaled negative log likelihood:
\begin{align*}
    E_{\theta}^{ClAM+CRP}(\vx) &\defeq -\frac{1}{\beta} \log \Bigg( \exp \Big(-(\beta^{-1} + \rho^{-1})^{-1}||\bm \vx||^2 \Big) (\alpha + K^+ d) \\
    &\quad\quad \quad\quad\quad\quad\quad  + \sum_{k=1}^{K^+} \exp \Big(-\beta ||\bm\mu_k - \vx||^2 \Big) (\tilde{\pi}_k - d) \Bigg)
\end{align*}

\begin{figure}
    \centering
    \includegraphics[width=0.8\textwidth]{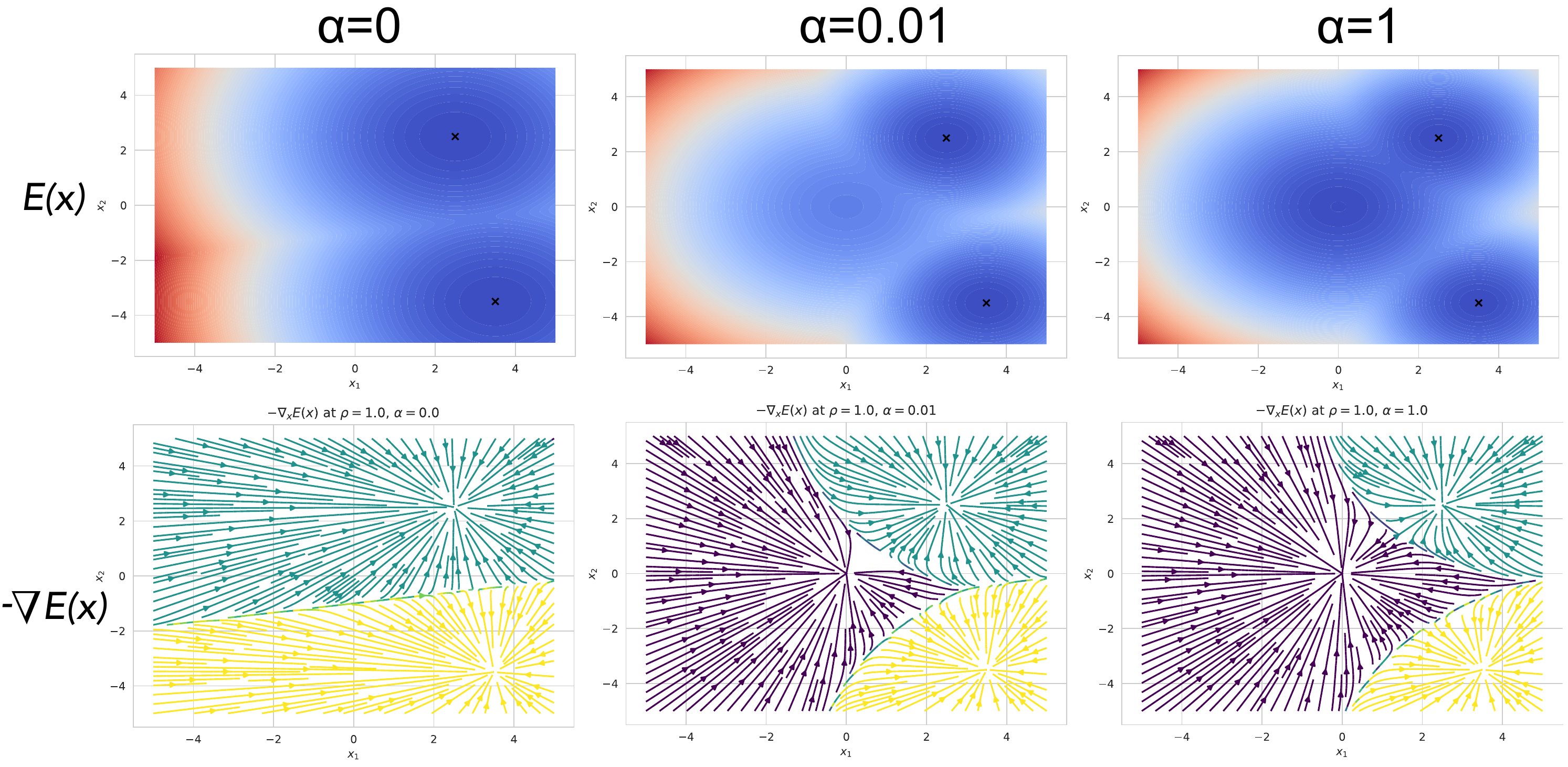}
    \vspace{2pt}
    \caption{\textbf{Energy landscape of new memory creation.} Left: Finite mixture models can result in each cluster's basin stretching out infinitely far. Middle and Right: Using the Chinese Restaurant Process, we endow the associative memory model with the ability to create new memories (cluster centroids) if the data is sufficiently far from existing memories: If a datum flows to the origin, we create a new memory for it. Hyperparameter $\alpha$ controls how likely new memories are to be created, with higher $\alpha$ attracting more points to the origin, causing faster cluster creation.}
    \label{fig:memory_creation_alpha}
\end{figure}

\paragraph{Nonparametric Latent Variable Energy Functions}
\label{subsec:nonparametric_latent_variable_energy_function}

One can then straightforwardly combine latent variable associative memory (Sec. \ref{subsec:latent_var_energy_functions}) with nonparametric associative memory (Sec. \ref{subsec:nonparametric_energy_functions}) to yield a nonparametric latent variable associative memory model:
\begin{equation}
    E_{\theta}^{ClAM+CRP+ELBO}(\vq) \defeq
    -\sum_{k=1}^K \vq_k \log p_{\theta}^{CRP}(\vx, z=k) + \sum_{k=1}^K \vq_k \log \vq_k.
    \label{eqn:ClAM_crp_elbo_energy}
\end{equation}

Interestingly, ClAM+CRP+ELBO shares some striking similarities with memory engrams \cite{josselyn2020memory}, an exciting new area of experimental neuroscience \cite{yiu2014neurons,rashid2016competition,park2016neuronal,lisman2018memory,pignatelli2019engram,lau2020role,jung2023examining} . 
Neurobiologically, we can view these dynamics as $K$ memory engrams that are self-excitatory and mutually inhibitory, with interactions given by $\diag(\vq) - \vq \vq^T$. We intend to explore this connection in subsequent work.

\section{Memory Capacity of Gaussian Kernel Density Estimators}

\begin{figure} 
    \centering
    \includegraphics[width=\linewidth]{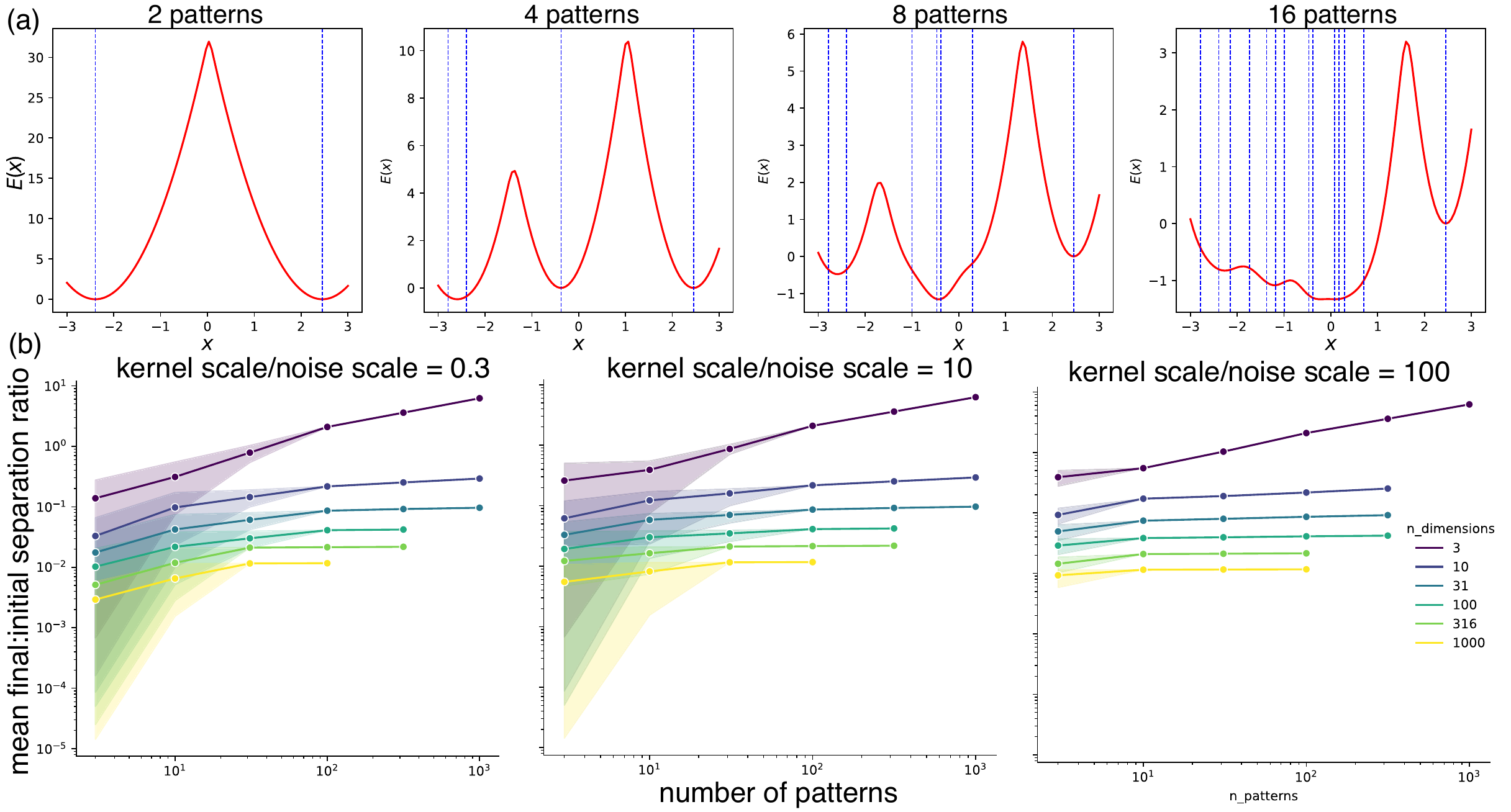}
    \caption{\textbf{KDE as associative memory: memory capacity limits.} (a) As more patterns are added, their energy basins (minima) merge, leaving us unable to retrieve them individually. (b) We quantify how well we can retrieve data by calculating the mean ratio of the distance between queries and their corresponding patterns after undergoing dynamics to before. We then normalize this ratio by the average distance of patterns. The smaller this ratio is, the closer the queries have converged to their corresponding patterns. We see that increasing the number of patterns results in poorer retrieval, while increasing the number of dimensions results in better retrieval.
    }\label{fig: kde_capacity_1d}
\end{figure}

An interesting problem commonly studied in the associative memory literature is analytically characterizing the memory retrieval, capacity, and failure behavior of memory systems \citep{gardner1988space, krotov2016dense,demircigil2017model, chaudhuri2019bipartite,lucibello2023exponential}.
In this section, we use such tools to study memory properties of kernel density estimators (KDEs), a widely used tool from probabilistic modeling \cite{parzen1962estimation,rosenblatt1956remarks,epanechnikov1969non,wand1994kernel,sheather1991reliable,hastie2009elements}.
Given $N$ i.i.d. samples $\mathcal{D} \defeq \{\vx_n\}_{n=1}^N \in \mathbb{R}^D$ from some unknown distribution, a kernel density estimator (KDE) estimates the unknown distribution as:
\begin{equation*}
    \hat{p}_{K,h}^{KDE}(\vx) \defeq \frac{1}{Nh}\sum_{n=1}^NK\Big(\frac{\vx-\vx_n}{h}\Big),
    \label{eqn:def_kernel_density_estimator}
\end{equation*}
with kernel function $K(\cdot)$ and bandwidth $h$.
The energy is defined as the negative log probability of the KDE:
\begin{equation} \label{eqn: KDE general energy}
    E_{K,h}^{KDE}(\vx) \defeq -\log \hat{p}_{K,h}^{KDE}(\vx) , %= - \log \Bigg( \sum_{n=1}^NK \Big(\frac{\vx-\vx_n}{h} \Big)\Bigg )+C,
\end{equation}
KDEs explicitly construct basin-like structures around each training datum, and thus can be viewed as memorizing the training data. We say that a pattern $\vx_n$ has been stored if there exists a ball with radius $R_n$, $S_n \defeq \{\vx \in \mathbb{R}^D : ||\vx - \vx_n||_2 \leq R_n \}$, centered at $\vx_n$ such that every point within $S_n$ converges to some fixed point $\vx_n^* \in S_n$ under the defined dynamics. The balls for different patterns must be disjoint.
We show here that KDEs have a finite memory storage and retrieval capacity  (Fig. \ref{fig: kde_capacity_1d}), by establishing a connection between the commonly used Gaussian KDE and the Modern Continuous Hopfield Network (MCHN) developed by Ramsauer et. al (2020)\citep{ramsauer2020hopfield}. This connection allows us to extend the capacity and convergence properties of the MCHN to the Gaussian KDE, showing that it has exponential storage capacity in the data dimensionality.
The widely used Gaussian KDE uses a Gaussian kernel with length scale (standard deviation) $\sigma$. Its energy is:
\begin{equation*}
    E_{\text{Gauss}, \sigma}(\vx) \defeq -\log\Bigg(\sum_{n=1}^N\exp\bigg(-\frac{1}{2\sigma^2}||\vx-\vx_n||^2\bigg)\Bigg).
    \label{eqn:kde gaussian energy}
\end{equation*}

% The dynamics of the Gaussian KDE are defined according to gradient descent on the energy landscape. For an arbitrary step size $\alpha$ we define the update rule: 
% %
% \begin{equation*}
%     \vx^{(i+1)} = 
%             \vx^{(i)} - \alpha \nabla E_{\text{Gauss}, \sigma}(\vx^{(i)}).
% \end{equation*}
%

In App. \ref{app:sec:gaussian_kde_capacity}, we prove that the energy and dynamics of the Gaussian KDE is exactly equivalent to the energy and update rule of the MCHN of Ramsaeuer et. al (2020) \citep{ramsauer2020hopfield}. Given the equivalence, we can characterize the capabilities and limitations of kernel density estimators in the same way as derived for MCHNs by Ramsaeuer et. al (2020)\citep{ramsauer2020hopfield}. Ergo, the capacity of the Gaussian KDE is shown to be: 
\begin{equation}
    C_{\text{Gauss}} = 2^{2(D-1)}.
\end{equation}
In Fig. \ref{fig: kde_capacity_1d}(b), we demonstrate numerically that Gaussian KDEs exhibit better retrieval at higher data dimensions and worse retrieval with more patterns.
\section{A Theoretical Justification for Pre-Normalization before Self-Attention}
\label{sec:transformer_normalization}

Next, we discover a way to understand the interaction between self-attention and normalization in transformers \cite{vaswani2017attention}.
The well-known equation for self-attention is:
\begin{equation*}
    \label{eqn:def_self_attention}
    \text{SA}(\mathbf{q},K,V) \defeq V \; \softmax \; (K \vq).
\end{equation*}
Previous work has connected self-attention to Hopfield networks \cite{ramsauer2020hopfield, millidge2022universal}.
However, transformers are not purely stacked self-attention layers; among many components, practitioners have found that applying normalization (e.g., LayerNorm \cite{ba2016layer}, RMS Norm \cite{zhang2019root}) \textit{before} self-attention significantly improves performance \cite{baevski2018adaptive, child2019generating, wang2019learning, xiong2020layer}. \textit{What effect does this composition of pre-normalization and self-attention have?} We show that the two together approximate clustering on the hypersphere using a mixture of inhomogeneous von Mises-Fisher (vMF) distributions \cite{fisher1953dispersion}. For concreteness, we consider LayerNorm, although RMS norm produces the same qualitative result.
%
%LayerNorm transforms a vector $\vx \in \mathbb{R}^D$ by computing its mean $\vm = (\sum_d \vx_d) / D$ and its variance $\sigma^2 = (\sum_d (\vx_d - \vm)^2)/D$, then shifting and scaling by learnable parameters $\bm \gamma, \bm \delta \in \mathbb{R}^D$:
%
\begin{equation*}
    \label{eqn:def_layer_normalization}
    LN_{\bm \gamma, \bm \delta}(\vx) \quad \defeq \quad \bm \gamma \odot \frac{\vx - \vm}{\sqrt{\sigma^2 + \epsilon}} + \bm \delta,
\end{equation*}
 where $\epsilon$ is a small constant for numerical stability and $\odot$ denotes elementwise multiplication. Recall that the vMF density function with unit vector $\vm_i \in \mathbb{R}^D, ||\vm_i||_2 
=1$ and concentration $\kappa_i \geq 0$ is:
 \begin{equation*}
    \label{eqn:def_vmf_density}
     p(\vx; \vm_i, \kappa_i) \propto \exp(\kappa_i \, \vm_i \cdot \vx).
 \end{equation*}

Let us define $\tilde{\vq}$ as the pre-shifted and scaled query i.e., $\vq \defeq \bm \gamma \odot \tilde{\vq} + \bm \delta$, with $||\tilde{\vq}||_2 \approx 1$. The $i^{\text{th}}$ element in the numerator of the softmax is:
\begin{align*}
    \exp(\vk_i \cdot \vq) = \exp(\vk_i \cdot (\bm \gamma \odot \Tilde{\vq} + \bm \delta))
    = \exp \Bigg(\underbrace{||(\vk_i \odot \bm \gamma)||_2}_{=\kappa_i} \underbrace{\frac{\vk_i \odot \bm \gamma}{||\vk_i \odot \bm \gamma||}}_{= \vm_i}\; \cdot\; \tilde{\vq} \Bigg) \; \underbrace{\exp \Bigg(\vk_i \cdot \bm \delta \Bigg)}_{= \pi_i}.
\end{align*}
Thus, LayerNorm followed by self-attention is equivalent to clustering with inhomogeneous vMF likelihoods and with (unnormalized) mixing proportions determined by the exponentiated inner products between the keys and the LayerNorm bias. A related commentary about the interaction between pre-LayerNorm and self-attention has been made before \cite{bricken2021attention}, albeit in a non-clustering and non-probabilistic context. This perspective suggests an unnecessary complexity exists in modern transformers between the keys $\{\vk_i \}$, scale $\bm \gamma$ and shift $\bm \delta$ in a way that might hamper expressivity. Specifically, if pre-LayerNorm composed with self-attention is indeed performing clustering, then each key $\vk_i$ is controlling both the concentration of the vMF likelihood as well as the mixing proportion $\pi_i$, and all keys must interact with the same scale $\bm \gamma$ and shift $\bm \delta$. Further, recent work has found that adding Layer Norm on the queries and keys stabilizes learning in ViTs \cite{dehghani2023scaling} and that this operation allows for training with large learning rates \citep{wortsman2023small} while avoiding instabilities \cite{zhai2023stabilizing}. 
Our proposed modification of the queries: $\vq \mapsto \bm \gamma \odot \Tilde{\vq} + \bm \delta$ indeed is equivalent to transforming $\vq \mapsto \text{LN}_{\gamma,\delta}(\vq) = \bm \gamma \odot \frac{\vq - \vm}{\sqrt{\sigma^2 + \epsilon}} + \bm \delta$.
% We propose fusing the two layers in the following manner: (1) keep the standardization of the LayerNorm input i.e. $(\vx - \vm) / \sqrt{\sigma^2 + \epsilon}$, (2) remove $\bm \gamma$, since this can be assimilated into each $\vk_i$ directly, and (3) replace $\vk_i \cdot \bm \delta$ with a learnable value $\delta_i \in \mathbb{R}$, which is an extra dimension sliced off from the key.

% We call this fused operation Self-Attention Layer-Normalization (SALN):
% %
% \begin{equation}
%     \text{SALN} \Big(\{\vk_i \}, \{\delta_i \}, \Tilde{\vq} \Big) \quad \defeq \quad  \text{Softmax}\Bigg(\vk_i \cdot \tilde{\vq} + \delta_i \Bigg),
% \end{equation}

% Unfortunately, this alternative layer does not save any parameters nor reduce the number of operations. Specifically, for $D$ dimensional vectors with batch size $B$ and sequence length $S$, then removing $\bm \gamma$ removes $O(D)$ parameters and $O(BSD)$ operations, but adding an extra dimension to the key adds $O(D)$ parameters to the key matrix and $O(BSD)$ operations to compute the $\{ \delta_i \}$.
% \section{Related Work}
% \label{sec:related_work}

% Many papers have commented on the relationship between associative memory models and modern advances in machine learning. \cite{radhakrishnan2018memorization, radhakrishnan2020overparameterized} \cite{ambrogioni2023search} \cite{ramsauer2020hopfield}

\section{Discussion}
\label{sec:discussion}

Associative memory and probabilistic modeling are two foundational fields of artificial intelligence that have remained (largely) unconnected for too long. While recent work has made good steps to demonstrate connections, e.g., to diffusion models \cite{ambrogioni2023search,hoover2023memory}, many more meaningful connections exist that our work hopefully demonstrates and inspires.

\clearpage
\nocite{schaeffer2023associative, schaeffer2021rcrp, schaeffer2022ribp, schaeffer2022dcrp}
\bibliographystyle{abbrvnat}
\bibliography{references_rylan}

% \clearpage

% \input{07_checklist}

\clearpage

%%%%%%%%%%%%%%%%%%%%%%%%%%%%%%%%%%%%%%%%%%%%%%%%%%%%%%%%%%%%%%%%%%%%%%%%%%%%%%%
%%%%%%%%%%%%%%%%%%%%%%%%%%%%%%%%%%%%%%%%%%%%%%%%%%%%%%%%%%%%%%%%%%%%%%%%%%%%%%%
% APPENDIX
%%%%%%%%%%%%%%%%%%%%%%%%%%%%%%%%%%%%%%%%%%%%%%%%%%%%%%%%%%%%%%%%%%%%%%%%%%%%%%%
%%%%%%%%%%%%%%%%%%%%%%%%%%%%%%%%%%%%%%%%%%%%%%%%%%%%%%%%%%%%%%%%%%%%%%%%%%%%%%%
\newpage
\appendix
\onecolumn

\section{Implementation Details for In-Context Learning of Energy Functions}
\label{app:sec:implementation_details_icl_ebm}

Our goal is to create new energy-based models that learn energy functions corresponding to conditional probability distributions without changing their parameters $\theta$. 
\begin{align}
    p_{\theta}^{ICL}(\vx|\mathcal{D}) = \frac{\exp\big( -E_{\theta}^{ICL}(\vx|\mathcal{D}) \Big)}{Z_{\theta}(\mathcal{D})}
\end{align}

To do this, we use causal GPT-style transformers \cite{vaswani2017attention, radford2018improving, radford2019language}. As background, in the context of conditional probabilistic modeling, a causal transformer is typically trained to output a conditional probability distribution at every index:
$$p(x_2|x_1), p(x_3|x_2, x_1), p(x_4|x_3,x_2, x_1), \dots$$

We simply replace each conditional distribution $p(x_n|x_{<n})$ with its corresponding energy function $E(x_n|x_{<n})$. This means that the transformer outputs a scalar variable at every index:
$$E(x_2|x_1), E(x_3|x_2, x_1), E(x_4|x_3,x_2, x_1), \dots$$

This scalar at each index is the model’s estimate of the energy at the last sample ($n^{\text{th}}$) input datum, based on an energy function constructed by the previous $n-1$ datapoints. The training pseudocode is:

\begin{lstlisting}[language=Python]
function training_step(batch, batch_idx):
    # Compute energy on real data.
    real_data = batch["real_data"]
    energy_on_real_data = transformer_ebm.forward(real_data)

    # Sample new confabulated data using Langevin MCMC.
    initial_sampled_data = batch["initial_sampled_data"]
    confab_data = sample_data_with_langevin_mcmc(real_data, initial_sampled_data)

    # Compute energy on sampled confabulatory data.
    energy_on_sampled_data = zeros(...)
    for seq_idx in range(max_seq_len):
        for conf_idx in range(n_confabulated_samples):
            real_data_up_to_seq_idx = clone(real_data[:, :seq_idx+1, :])
            real_data_up_to_seq_idx[:, -1, :] = sampled_data[:, conf_idx, seq_idx, :]
            energy_on_confab_data = transformer_ebm.forward(real_data_up_to_seq_idx)
            energy_on_sampled_data[:, conf_idx, seq_idx, :] += energy_on_confab_data[:, -1, :]

    # Compute difference in energy between real and confabulatory data. 
    diff_of_energy = energy_on_real_data - energy_on_sampled_data

    # Compute total loss.
    total_loss = mean(diff_of_energy)

    return total_loss
\end{lstlisting}

\clearpage
\section{Experiment Details for Latent Variable \& Bayesian Nonparametric Associative Memory Models}
\label{app:sec:implementation_details_latent_var_bayesian_nonparam_am}

For our clustering experiments (Sec. \ref{sec:learning_memories_for_associative_memory}), we largely follow the experimental setup established by \citet{saha2023end}, but make key modifications. We consider the same datasets largely taken from the UCI Machine Learning Repository \cite{UCI-ML-Repo}
(Table \ref{app:tab:clustering_dataset_info}): 

\begin{table}[ht]
\centering
\begin{tabular}{|l|c|c|c|c|}
\hline
Dataset Name & Year & Num Samples & Num Features & Num Classes \\
\hline
Boston Housing & 1993 & 506 & 14 & 46 \\
Wisconsin Breast Cancer & 1995 & 569 & 30 & 2 \\
Ecoli & 1996 & 336 & 7 & 8 \\
Fashion MNIST & 2017 & 60000 & 784 & 10 \\
Forest Covertype & 1998 & 581012 & 54 & 7 \\
Image Segmentation & 1990 & 2310 & 19 & 7 \\
Iris & 1936 & 150 & 4 & 3 \\
KDD Cup & 1999 & 494021 & 38 & 23 \\
Libras Movement & 2009 & 360 & 90 & 15 \\
Microarray GCM & 2001 & 190 & 16063 & 14 \\
Olivetti Faces & 1992 & 400 & 4096 & 40 \\
USPS Digits & 1994 & 9298 & 256 & 10 \\
Wine & 1994 & 178 & 13 & 3 \\
Yale Faces & 1997 & 165 & 1024 & 15 \\
Zoo & 1990 & 100 & 16 & 7 \\
\hline
\end{tabular}
\caption{\textbf{Summary of Datasets for Clustering Experiments.}}
\label{app:tab:clustering_dataset_info}
\end{table}

For metrics, we considered 4 supervised metrics (Rand Score, Adjusted Rand Score, Adjusted Mutual Info Score, Normalized Mutual Info Score) and 3 unsupervised metrics (Calinski-Harabasz Score, Davies-Bouldin Score, Silhouette Score). We chose to use multiple metrics because different metrics are known to have different trade-offs and we wanted to make clear that we did not cherrypick a particular metric that favored our results. 

For each clustering algorithm, we chose hyperparameters to (1) be reasonable, (2) be relatively diverse and (3) yield approximately the same number of clustering fits as all the other models. We include the hyperparameter sweeps for each method below:

\clearpage
\section{Capacity, Retrieval Errors and Memory Cliffs of Gaussian Kernel Density Estimators}
\label{app:sec:gaussian_kde_capacity}

We characterize the capacity and memory cliffs of kernel density estimators, i.e. how much data can be successfully retrieved by following the negative gradient of the log probability, and what happens when that limit is exceeded? Suppose we have $N$ training data $\{\vx_n\}_{n=1}^N \in \mathbb{R}^D$, and we consider the estimated probability distribution by a kernel density estimator (KDE):
\begin{equation}
    \hat{p}_{K,h}(\vx) \defeq \frac{1}{Nh}\sum_{n=1}^NK\Big(\frac{\vx-\vx_n}{h}\Big),
\end{equation}
with kernel function $K(\cdot)$ and bandwidth $h$. The energy is defined as the negative log probability of the KDE:
\begin{equation} \label{eqn: KDE general energy}
    E_{K,h}(\vx) \defeq -\log(\hat{p}_{K,h}(\vx)) = - \log \Bigg( \sum_{n=1}^NK \Big(\frac{\vx-\vx_n}{h} \Big)\Bigg )+C,
\end{equation}
where $C$ is a constant that will not affect dynamics and will be omitted moving forward. %This is analogous to uniformly adjusting the energy landscape by a constant amount, which does not alter the energy gradient responsible for driving the dynamics.
To characterize the capacity and failure modes of kernel density estimators, we begin with relevant definitions (many from \cite{ramsauer2020hopfield}).

\begin{definition}[Separation of Patterns]
The separation $\Delta_n$ of a pattern (i.e. a training datum) $\vx_n$ from the other patterns is defined as one-half the squared distance to the closest training datum:
\begin{equation*}
    \Delta_n \defeq \frac{1}{2} \cdot \min_{n' \neq n} ||\vx_n-\vx_{n'}||^2.
\end{equation*}
\label{def:separation}
\newline
\end{definition}

\begin{definition}[Pattern Storage]
We say that a pattern $\vx_n$ has been stored if there exists a ball with radius $R_n$, $S_n \defeq \{\vx \in \mathbb{R}^D : ||\vx - \vx_n||_2 \leq R_n \}$, centered at $\vx_n$ such that every point within $S_n$ converges to some fixed point $\vx_n^* \in S_n$ under the defined dynamics. This point $\vx_n^*$ is not necessarily the training point $\vx_n$. 
The balls associated with different patterns must be disjoint, i.e. $\forall n' \neq n: S_{n'} \cap S_n = \emptyset$. The value $R_n$ is called the radius of convergence.
\label{def:storage_and_retrieval}
\newline
\end{definition}

\begin{definition}[Retrieval Error]
    For a stored pattern $\vx_n$, let $S_n$ be the ball around $\vx_n$ as defined in \ref{def:storage_and_retrieval}. By definition \ref{def:storage_and_retrieval}, every point within the $S_n$ must converge to some $\vx_n^*$. We define the \textbf{retrieval error} to be $||\vx_n-\vx_n^*||$.
    
\end{definition}

\begin{definition}[Storage Capacity]
The storage capacity of a particular associative memory model is the number of patterns $C$ such that all $C$ patterns $\vx_1, ..., \vx_C$ are stored under Def. \ref{def:storage_and_retrieval}.
\label{def:storage_capacity}
\newline
\end{definition}

\begin{definition}[Largest Norm of Training Data]
    We define $M$ as the largest $L^2$ norm of our training data:
    \begin{equation*}
        M = \max_n ||\vx_n||_2.
    \end{equation*}
\end{definition}

\subsection{Kernel Density Estimator with a Gaussian Kernel}

We begin by studying the widely used Gaussian KDE with length scale (standard deviation) $\sigma$. Its energy function is:
\begin{equation}
    E_{\text{Gauss}, \sigma}(\vx) \defeq -\log\Bigg(\sum_{n=1}^N\exp\bigg(-\frac{1}{2\sigma^2}||\vx-\vx_n||^2\bigg)\Bigg).
    \label{eqn:kde gaussian energy}
\end{equation}

To study the capacity, retrieval error and memory cliff of the Gaussian KDE, it will be helpful to briefly summarize the modern continuous  Hopfield network (MCHN) of \citet{ramsauer2020hopfield}.

\begin{definition}[MCHN Energy Function]
    The MCHN energy function is given as 
    \begin{equation}
        E_{\text{MCHN}}(\vx) \defeq -\beta^{-1} \log \Bigg(\sum_{n=1}^N\exp\bigg(\beta \vx_n^T\vx \bigg)\Bigg) + \beta^{-1}\log(N) +\frac{1}{2}\vx^T\vx+\frac{1}{2}M^2
    \end{equation}
    where $\beta$ is the inverse temperature. 
    \newline
    \label{eqn: original MCHN energy}
\end{definition}

\begin{definition}[MCHN Dynamics]
    Defining the matrix $\rmX$ whose columns are our training points $\vx_n$:
    $$\rmX \defeq \begin{bmatrix}
    \vline \;\;\;\; \vline \;\;\;\; \;\;\;\; \;\;\;\; \vline \\
        \vx_1 \; \vx_2 \; \dots \; \vx_N \\
        \vline \;\;\;\; \vline \;\;\;\; \;\;\;\; \;\;\;\; \vline
    \end{bmatrix},\in \mathbb{R}^{D \times N},$$
    the update rule introduced by \citet{ramsauer2020hopfield} is defined to be $$\vx^{(i+1)} = \rmX \softmax\bigg(\beta\rmX^T\vx^{(i)}\bigg),$$
    which corresponds to the Concave-Convex Procedure (CCCP) for minimizing the energy function in \ref{eqn: original MCHN energy} 
    \newline
    \label{eqn: MCHN dynamics}
\end{definition}

To calculate the convergence and capacity properties of the MCHN, \citet{ramsauer2020hopfield} assume that all the training points lie on a sphere. 
\begin{assumption}[All training points lie on a sphere]
\label{asmptn: all pts on sphere}
    Recall that $M$ is defined as the largest norm of our training data. Moving forward, we assume that the points $\vx_1, ..., \vx_N$ are distributed over a sphere of radius $M$, i.e. that
\begin{equation*}
    ||\vx_1||= \dots =|| \vx_N|| = M.
\end{equation*}
\end{assumption}

Next, we will show that under assumption \ref{asmptn: all pts on sphere}, the Gaussian KDE has identical energy and dynamics to the MCHN. Consequently, we are able to extend the capacity and convergence properties of the MCHN derived by \citet{ramsauer2020hopfield} to the Gaussian KDE, showing that it has exponential storage capacity in $D$, the number of dimensions of our data.

\begin{theorem}
    The Gaussian KDE energy function is equivalent to the MCHN energy function.
\end{theorem}

\begin{proof}
    We begin by simplifying the MCHN energy equation in \ref{eqn: original MCHN energy}. We have 
    \begin{equation*}
    \begin{split}
        E_{\text{MCHN}}(\vx) &= -\beta^{-1} \log \Bigg(\sum_{n=1}^N\exp\bigg(\beta \vx_n^T\vx \bigg)\Bigg) + \beta^{-1}\log(N) +\frac{1}{2}\vx^T\vx+\frac{1}{2}M^2 \\
        &= -\beta^{-1} \log \Bigg(\sum_{n=1}^N\exp\bigg(-\frac{1}{2}\beta\Big(M^2-||\vx_n||^2\Big)\bigg)\exp\bigg(-\frac{1}{2}\beta||\vx-\vx_n||^2\bigg)\Bigg) + \beta^{-1}\log(N).
    \end{split}
    \end{equation*}
    Under assumption \ref{asmptn: all pts on sphere}, and using inverse temperature $\beta = \frac{1}{\sigma^2}$, we can further simplify this equation to get
    \begin{equation}
        E_{\text{MCHN}}(\vx) = -\sigma^2 \log \Bigg(\sum_{n=1}^N\exp\bigg(-\frac{1}{2\sigma^2}||\vx-\vx_n||^2\bigg)\Bigg) + \sigma^2\log(N),
    \end{equation}
    which is a scaled and shifted version of the energy function in \ref{eqn:kde gaussian energy}. Ergo, the Gaussian KDE energy function is equivalent to the MCHN energy function.
\end{proof}

\begin{theorem}
    The Gaussian KDE with appropriate step size has identical dynamics to the MCHN.
\end{theorem}

\begin{proof}
    For the Gaussian KDE in \ref{eqn:kde gaussian energy}, the dynamics are defined by gradient descent on the energy landscape with step size $\alpha$:
    \begin{equation} \label{eqn: general kde update}
        \begin{split}
            \vx^{(i+1)} &= 
            \vx^{(i)} - \alpha \nabla E_{\text{Gauss}, \sigma}(\vx^{(i)}) \\
            &= \vx^{(i)}-\frac{\alpha} {\sigma^2}\cdot \frac{\sum_{n=1}^N \exp\bigg(-\frac{1}{2\sigma^2}||\vx^{(i)}-\vx_n||^2\bigg)(\vx^{(i)}-\vx_n)}{\sum_{n=1}^N \exp\bigg(-\frac{1}{2\sigma^2}||\vx^{(i)}-\vx_n||^2\bigg)}.
        \end{split}
    \end{equation}
    Using the assumption \ref{asmptn: all pts on sphere}, we can further simplify the exponent to get
    \begin{equation*}
        ||\vx^{(i)}-\vx_n||^2 = ||\vx^{(i)}||^2+M^2-2\vx_n^T\vx^{(i)}.
    \end{equation*}
    Substituting in \ref{eqn: general kde update}, and canceling out the common factors we get:
    \begin{equation*}
        \vx^{(i+1)} = \vx^{(i)}-\frac{\alpha} {\sigma^2}\cdot \frac{\sum_{n=1}^N \exp\bigg(\frac{1}{\sigma^2}\vx_n^T\vx^{(i)}\bigg)(\vx^{(i)}-\vx_n)}{\sum_{n=1}^N \exp\bigg(\frac{1}{\sigma^2}\vx_n^T\vx^{(i)}\bigg)}.
    \end{equation*}
    Choosing step size $\alpha=\sigma^2$, we get the update rule:
    \begin{equation} \label{eqn:Gaussian kde update rule}
    \begin{split}
        \vx^{(i+1)} 
        &= \vx^{(i)}-\vx^{(i)}+\frac{\sum_{n=1}^N \vx_n\exp\bigg(\frac{1}{\sigma^2}\vx_n^T\vx^{(i)}\bigg)}{\sum_{n=1}^N \exp\bigg(\frac{1}{\sigma^2}\vx_n^T\vx^{(i)}\bigg)} \\
        &= \sum_{n=1}^N \vx_n \softmax\bigg(\frac{1}{\sigma^2}\vx_n^T\vx^{(i)}\bigg) \\ 
        &= \rmX \softmax\bigg(\beta\rmX^T\vx^{(i)}\bigg),
    \end{split}
    \end{equation}
    which is precisely the update rule described in \ref{eqn: MCHN dynamics}.    
\end{proof}

We have demonstrated an equivalence between the energy functions and update rules of MCHNs and Gaussian KDEs. We now apply convergence and storage capacity analysis for MCHNs to Gaussian KDEs \citet{ramsauer2020hopfield}.

%TODO: interpret results in English. explain where sigma comes in and consider cases
\begin{proposition}
If the training points are well separated, the Gaussian KDE has a radius of convergence equal to $\frac{\sigma^2}{NM}$.
\end{proposition}
\begin{proof}
    We assume that the data $\vx_n$ is well-separated. Concretely, we have: 
    \begin{assumption}[Well-Separated Data]\label{asmptn: well separated data}
        \begin{equation}
            \Delta_n \geq \frac{2\sigma^2}{N}+\sigma^2\log\bigg(\frac{2}{\sigma^2}(N-1)NM^2\bigg).
        \end{equation}
    \end{assumption} 
    Defining the ball around $\vx_n$:
    \begin{equation*}
        S_n \defeq \left\{\vx \;\middle| \; \; ||\vx-\vx_n|| \leq \frac{\sigma^2}{NM}\right\},
    \end{equation*}

    \citet{ramsauer2020hopfield}
    show that our update rule in \ref{eqn:Gaussian kde update rule} is a contraction mapping over the ball $S_n$. Thus, by Banach's fixed point theorem, the update rule converges to a fixed point within the ball after sufficient iterations. Thus, by our definition of storage and retrieval, the point $\vx_n$ will be stored and the radius of $S_n$ gives the radius of convergence:
    \begin{equation}
        R_n = \frac{\sigma^2}{NM}.
    \end{equation}
\end{proof}

\begin{figure} 
    \centering
    \includegraphics[width=\linewidth]{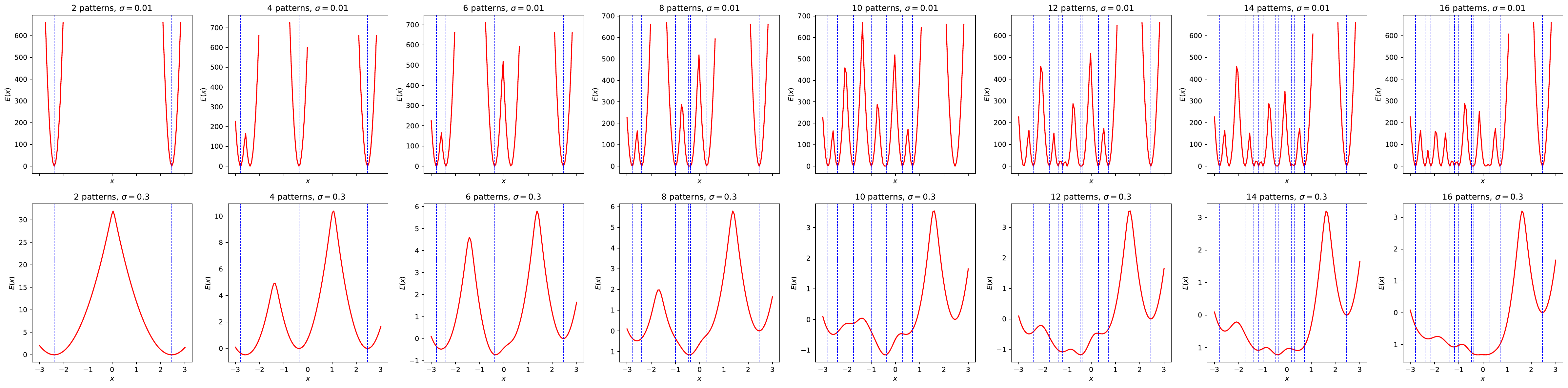}
    \caption{\textbf{Energy landscape under different numbers of patterns, with two different standard deviation values $\sigma$.} The basins for different patterns are more likely to merge when the Gaussian has a larger standard deviation, and when the patterns are too close together. The latter is likely to happen when we attempt to store too many patterns in a finite space. When two basins merge, we are unable to retrieve the corresponding patterns individually.
    }\label{fig: kde_capacity_1d_standard_dev}
\end{figure}

Intuitively, if our patterns get too close, their corresponding basins in the energy function merge, leaving us unable to retrieve either of them individually. This can be seen in the lower panel of Fig. \ref{fig: kde_capacity_1d_standard_dev} 
\\\\
Assumption \ref{asmptn: well separated data} establishes a lower bound for just how close the patterns can get without their energy basins merging. This depends on the standard deviation of the Gaussian, number of data points, and the radius of the sphere they are distributed over. Intuitively, if the standard deviation of the Gaussian is large, the basins are more likely to merge, and thus the lower bound for $\Delta_n$ increases with $\sigma$. In Fig. \ref{fig: kde_capacity_1d_standard_dev}, we can observe the effects of $\sigma$ on the energy landscape. A smaller $\sigma$ allows for patterns to be closer before their basins merge.
\\\\
Additionally, notice that for large $N$ (meaning that we have a lot of training points), the lower bound for $\Delta_n$ increases with $N$, signifying the fact that the dynamics near each basin can be overwhelmed by the collective effects of multiple other basins. Therefore, the more training points we have, the more we need to separate out the training points in order to safely retrieve them.
\\\\
Now, we turn our attention to the storage capacity of the Gaussian KDE.
\begin{proposition} \label{gaussian kde capacity}
    If the training points are sufficiently well-separated and we have $M = 2\sqrt{D-1}$ and $D\geq 4$, or $M = 1.7\sqrt{D-1}$ and $D \geq 50$, the Gaussian KDE can store exponentially many patterns in $D$, the dimensions of the data.
\end{proposition}

\begin{proof}
    We assume that the patterns are spread equidistantly over a sphere of radius $M$, and take $\sigma = 1$. The patterns are assumed to be well separated so that 
\begin{equation*}
    \Delta_{min} \geq \frac{2\sigma^2}{N}+\sigma^2\log\bigg(\frac{2}{\sigma^2}N^2M^2\bigg).
\end{equation*}
Under these conditions, \cite{ramsauer2020hopfield} show that at least 
$$N = 2^{2(D-1)}$$
patterns can be stored, so the storage capacity of the Gaussian KDE is $C_{\text{Gauss}} = 2^{2(D-1)}$.
\end{proof}

A more thorough analysis of storage capacity under different assumptions (such as for randomly placed patterns) can be found in \citet{ramsauer2020hopfield}.

\begin{figure} 
    \centering
    \includegraphics[width=\linewidth]{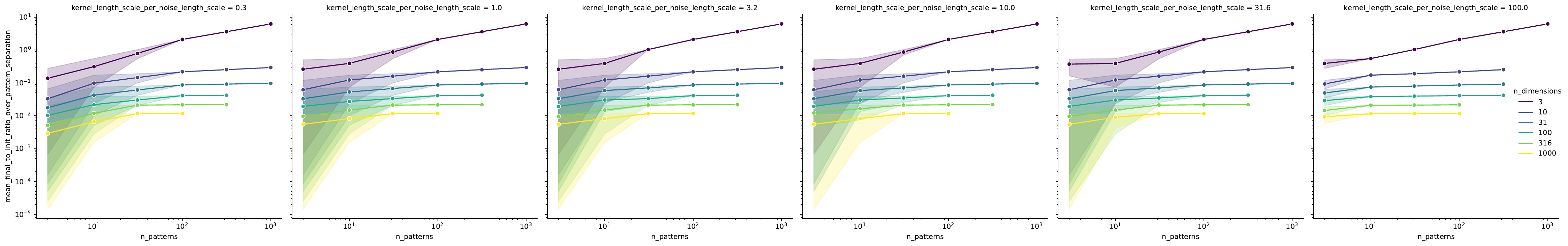}
    \caption{\textbf{KDE as associative memory: memory capacity limits.} We sample $N$ patterns on a $D$-dimensional hypersphere of radius $M=2\sqrt{D-1}$, which we use to define our energy landscape. We then initialize 100 particles perturbed from the positions of each pattern, and let them evolve under the energy function. We calculate the mean ratio of the distance between particles and their corresponding patterns after undergoing dynamics, divided by this distance at initialization. We then normalize this ratio by the average distance of patterns. The smaller this ratio is, the closer the particles have converged to their corresponding patterns. We see that increasing the number of patterns results in poorer retrieval, while increasing the number of dimensions results in better retrieval.
    }
\end{figure}

% \subsection{KDE with an Epanechnikov kernel} 
% \begin{definition}\label{def: Epanechnikov kernel}
%     The Epanechnikov kernel: This kernel is defined as $$K(\vx) = \frac{3}{4}(1-||\vx||^2) \;\; \text{ with $||\vx||\leq 1.$}$$
% \end{definition}

% \begin{definition}
%     At any given point $\vx$, the number of training points within radius $h$ of $\vx$ is defined to be $N^*$. TODO:
% \end{definition}

% \begin{definition} \label{Epanechnikov KDE energy}
%     Plugging the Epanechnikov kernel into equation \ref{eqn: KDE general energy}, we define the energy function of the Epanechnikov KDE:
% %
% \begin{equation}
%     E_{\text{Ep,$h$}}(\vx) \defeq -\log\Bigg(\sum_{n=1}^{N^*}\bigg(1-\frac{||\vx-\vx_n||^2}{h^2}\bigg)\Bigg),
% \end{equation}
% %
% where $N^*$ is defined as the number of training points $\vx_n$ within radius $h$ of $\vx$. These are the only training points that contribute to the energy, since the input of the Epanechnikov kernel must have L-2 norm smaller than 1. TODO: explain this better
% \end{definition}

\end{document}